\documentclass[sigplan,screen,nonacm]{acmart}
\usepackage{amsmath}
\AtBeginDocument{%
  \providecommand\BibTeX{{%
    \normalfont B\kern-0.5em{\scshape i\kern-0.25em b}\kern-0.8em\TeX}}}

\begin{document}

\title{Gemini Pro Defeated by GPT-4V: Evidence from Education
}


\author{Gyeong-Geon Lee}

\affiliation{%
  \institution{AI4STEM Education Center, University of Georgia}
  \institution{Department of Mathematics, Science and Social Studies Education, University of Georgia}
  \city{Athens}
  \state{Georgia}
  \country{USA}
  \postcode{30602}
}
\orcid{1234-5678-9012}

\author{Ehsan Latif}

\affiliation{%
  \institution{AI4STEM Education Center, University of Georgia}
  \institution{Department of Mathematics, Science and Social Studies Education, University of Georgia}
  \city{Athens}
  \state{Georgia}
  \country{USA}
  \postcode{30602}
}
\orcid{1234-5678-9012}

\author{Lehong Shi}

\affiliation{%
  \institution{Department of Workforce Education and Instructional Technology, University of Georgia}
  \city{Athens}
  \state{Georgia}
  \country{USA}
  \postcode{30605}
}
\orcid{1234-5678-9012}

\author{Xiaoming Zhai}

\authornote{This is the corresponding author: xiaoming.zhai@uga.edu}
\affiliation{%
  \institution{AI4STEM Education Center, University of Georgia}
  \institution{Department of Mathematics, Science and Social Studies Education, University of Georgia}
  \city{Athens}
  \state{Georgia}
  \country{USA}
  \postcode{30602}
}
\orcid{1234-5678-9012}

\renewcommand{\shortauthors}{Lee, Latif, Shi, Zhai}

\begin{abstract}
  This study compared the classification performance of Gemini Pro and GPT-4V in educational settings. Employing visual question answering (VQA) techniques, the study examined both models' abilities to read text-based rubrics and then automatically score student-drawn models in science education. We employed both quantitative and qualitative analyses using a dataset derived from student-drawn scientific models and employing NERIF (Notation-Enhanced Rubrics for Image Feedback) prompting methods. The findings reveal that GPT-4V significantly outperforms Gemini Pro in terms of scoring accuracy and Quadratic Weighted Kappa. The qualitative analysis reveals that the differences may be due to the models' ability to process fine-grained texts in images and overall image classification performance. Even adapting the NERIF approach by further de-sizing the input images, Gemini Pro seems not able to perform as well as GPT-4V. The findings suggest GPT-4V's superior capability in handling complex multimodal educational tasks. The study concludes that while both models represent advancements in AI, GPT-4V's higher performance makes it a more suitable tool for educational applications involving multimodal data interpretation.
\end{abstract}







\settopmatter{printfolios=true}
\maketitle

\section{Introduction}

The keynote and education methods have been in step with the concurrent technologies \citep{oecd2019learning}. Nowadays, artificial intelligence (AI) technologies based on deep neural networks are getting closer to realizing the long-pursued AI in education (AIEd) initiative \citep{luckin2016intelligence, hwang2020vision}. Especially in the 2020s, the development and release of large language models (LLMs), such as BERT (Bidirectional Encoder Representations from Transformers) and GPT (Generative Pre-trained Transformer) are facilitating the integration of AI technologies into educational research and practice due to their high capabilities of natural language processing, understanding and reasoning, and generation for various tasks such as teaching material preparation, data augmentation, item generation, and automatic scoring \citep{lo2023impact, fang2023using, lee2023fewshot, latif2023finetuning}. Particularly, user-friendly LLMs such as ChatGPT have hugely impacted the global trend of AI usage in research and everyday lives by enabling conversational interactions between humans and machines based on natural languages.

What even sputters the innovations in AIEd research is the realization of visual question answering (VQA), which enables machines to answer the user's natural language-based query on a given image \citep{antol2015vqa}, which can assist in learning and teaching. VQA has the potential in educational settings when using visual data to interpret students' ideas on the learning content or classroom dynamics \citep{lee2023multimodality}. For example, teachers may feed a computer with a student-drawn model to explain refraction, and the computer can recognize the lights and angles and provide feedback for improvements\citep{RN3371}. VQA, in principle, suggests the possibility of multimodal interaction between humans and computers, enabling multimodal learning and assessment with AI \citep{lee2023multimodality}. 

Nevertheless, there have been scarce studies that employed VQA techniques in educational contexts. One of the reasons for this gap could be found in a barrier for general educational scholars, who might not have access to the essential technologies and resources. Therefore, we can deduce that if general users could utilize VQA techniques as they use LLMs via a chat-like interface, it would greatly help synergize cutting-edge technologies and education. When VQA is attached to LLMs, the models are called vision-language models (VLMs). 

GPT-4V, released on September 2023 by \citeN{openai2023gptsee}, has provided the global users with the affordance of VLM, as an extended module of ChatGPT and GPT-4. Since their release, ChatGPT and GPT-4 have notably influenced educational research and applications due to the high capability of natural language processing, understanding, and generation for various educational tasks and applications. Much research \citep{dempere2023impact, lo2023impact, latif2023finetuning, ausat2023can} suggested the effectiveness of ChatGPT in various educational tasks, such as aiding automated assessment, suggesting educational materials, and facilitating personalized learning. For example, \citeN{zhai2022chatgpt} conducted a piloting study of user experiences, focusing on the effectiveness of using ChatGPT as a tool to support writing a research paper, suggesting its potential and benefit in processing natural language-based tasks. However, the recent release of GPT-4V expanded its educational potential to facilitate learning and assessments featuring multimodalities, such as text and image \citep{lee2023nerif, lee2023multimodality}. For example, \citeN{lee2023nerif} tested GPT-4V's performance of automatic scoring student-drawn models by processing problem image and textual context with rubric and reported its multinomial classification accuracy as mean =.51, SD = .037. While this level of accuracy is remarkable, effort is needed to improve it before this method can be used in classrooms. 

The release of Gemini Pro, an extended module of Bard, by Google DeepMind in December 2023 provided an emergent opportunity to fill this gap. Google DeepMind claimed that Gemini "is built from the ground up for multimodality - reasoning seamlessly across text, images, video, audio, and code" and is the first model to outperform human experts on MMLU (massive multitask language understanding) since it \citep{deepmind_gemini}.
Google Bard and Gemini Pro were released only a few months later than ChatGPT and GPT-4V, which shows the technical competition for state-of-the-art AI services. Google Bard and Gemini Pro have also been tested and used for reasoning, answering knowledge-based questions, solving math problems, translating between languages, generating code, and acting as instruction-following agents through  benchmarks \cite{gemini}. \citeN{akter2023depth} performed an extensive study to test the capabilities and functionalities of Gemini and compared it against GPT-3.5 turbo and found that Gemini underperforms GPT-3.5 turbo in reasoning, generating code, and solving math problems. \citeN{waisberg2023google} provided a side-by-side comparison of Bard (underlying Gemini-pro) with ChatGPT for its application in ophthalmology. \citeN{mcintosh2023google} provided a comprehensive survey on their transformative impacts of Mixture of Experts (MoE), multimodal learning and suggested the speculated advancements towards Artificial General Intelligence  \cite{goertzel2014artificial,Latif2023AGI}. Furthermore, Microsoft researchers \cite{liu2023evaluation} performed comprehensive evaluation of the performances of GPT-4V and Gemini Pro using VQA online dataset \cite{chen2023fully} and reported that the average accuracy of GPT-4V and Gemini is 0.53 and 0.42, respectively. If Gemini Pro's performance is also promising for educational tasks, it could be used for a variety of purposes while competing with GPT-4V. Also, since there have been scarce studies on the use of Gemini Pro for education, exploring prompt engineering methods to elicit its full potential will also contribute to the AIEd research field.

However, as known to the authors, no studies have examined the eligibility of Gemini Pro in educational settings, not to mention the comparison of its performance with GPT-4V in dealing with VQA tasks. In this situation, with competing technologies struggling for SOTA, we suggest that comparing their applicability and effectiveness to educational studies is of timely need and could give a direction to the ongoing AIEd initiative regarding which VLM model could best serve educational tasks for researchers and practitioners. In this study, we answered to the following research questions.
1. How are the scoring performance of GPT-4V and Gemini Pro?
2. What are the characteristics of GPT-4V and Gemini Pro? 
3. How Gemini Pro's performance on an educational task can be improved?

\section{Gemini Pro vs GPT-4V}\label{comparison_review} 
The development of Large Multimodal Models (LMMs) aims to expand upon the capabilities of Large Language Models (LLMs) by integrating multi-sensory skills to achieve even stronger general intelligence, thus supporting more natural human-computer interactions \cite{yang2023dawn}. LMMs have one shared embedding space to integrate and process multiple data modalities, such as text, image, audio, video, 3D, etc. GPT-4V and Gemini Pro, the latest LMMs, can take images and/or text as inputs to perform various language, vision, and vision-language tasks, such as language translation and coding \cite{devlin2018bert}, image recognition \cite{yu2023applications}, object localization \cite{yang2023dawn}, visual question answering \cite{li2023comprehensive}, and visual dialogue \cite{zhu2023minigpt}. Furthermore, in the context of education, LLMs can be used for automatic scoring \citep{latif2023automatic} by applying chain-of-thought \citep{lee2023applying}.  A close examination of the two state-of-the-art models in terms of their architecture, training approach, training datasets, performance, capabilities, safety, and applications can illuminate the usability and affordances of the two models, especially in education.

\subsection{Architecture}
\begin{figure*}
    \centering
    \includegraphics[width=1\linewidth]{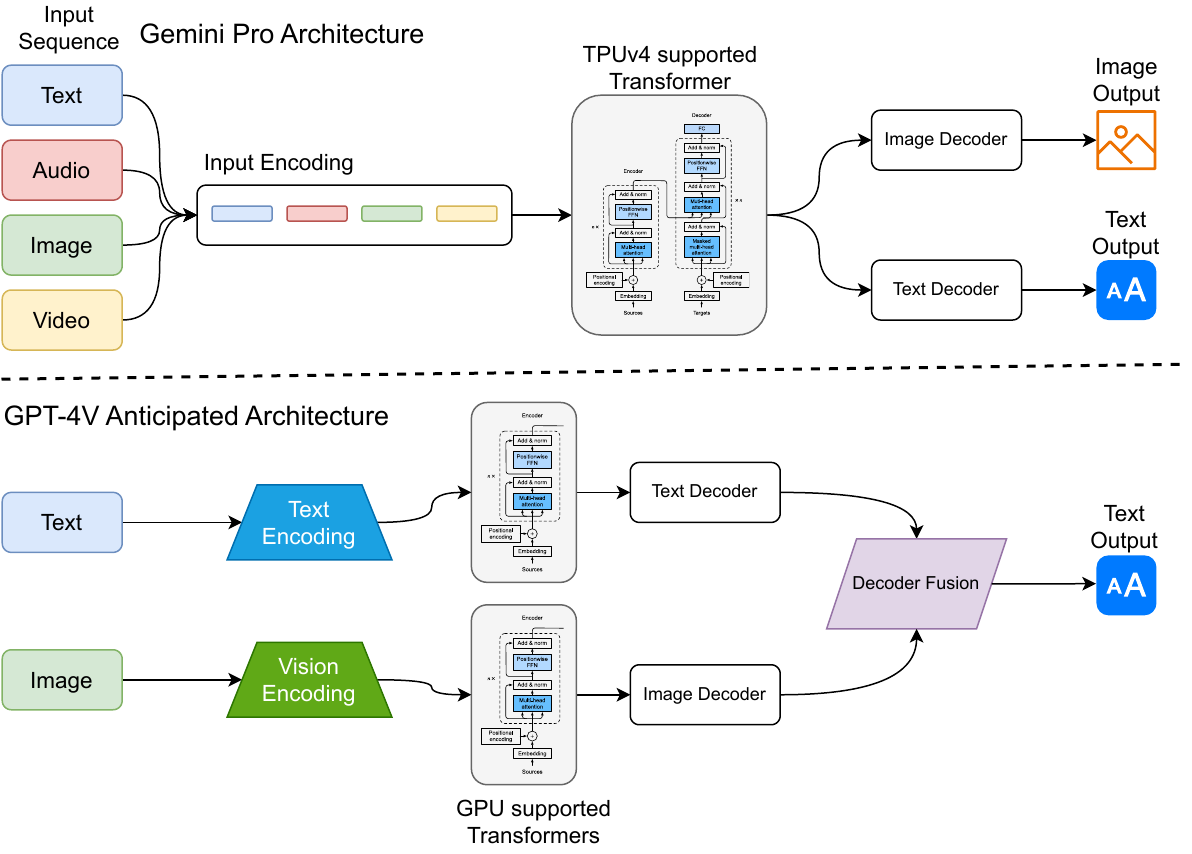}
    \caption{Architectural Comparison of Gemini Pro and GPT-4V}
    \label{fig:architecture}
\end{figure*}

GPT-4V, a product of OpenAI, is built on a transformer-based model \cite{vaswani2017attention} designed to understand context and meaning through relationships in sequential data. This architecture is known for its ability to handle complex language and image-processing tasks \cite{gpt4technical}. In contrast, Gemini Pro, developed by Google, is a large multimodal model that goes beyond processing text and images to include audio and video inputs. This broader range of input types suggests a more versatile and comprehensive approach to multimodal learning \cite{gemini}. The architectural comparison can be seen in Fig.~\ref{fig:architecture}, as OpenAI has not released the exact architecture, we have anticipated its architecture based on the available information\cite{gpt4visiontechnical}.

\subsection{Training Approach}
The training regime of GPT-4V includes advanced techniques such as Reinforcement Learning from Human Feedback (RLHF) \cite{christiano2017deep}. This approach refines the model's output based on human input, ensuring more accurate and contextually relevant responses. Additionally, GPT-4V employs a loss prediction method grounded in Power Law \cite{kaplan2020scaling}, utilizing scaling laws to optimize training efficiency \cite{henighan2020scaling}. On the other hand, Gemini Pro is trained using Google's tensor processing unit TPUv4 \cite{jouppi2023tpu}, which offers high computational power. Its extended 32K context length allows for the processing of much larger chunks of data at once, potentially leading to deeper insights and understanding in complex tasks \cite{gemini}. TPUs are trained to support 32k context length, employing efficient attention mechanisms (e.g. multi-query attention \cite{shazeer2019fast}). TPUv4 accelerators are deployed in “SuperPods” of 4096 chips, each connected to a dedicated optical switch, which can dynamically reconfigure 4x4x4 chip cubes into arbitrary 3D torus topologies in around 10 seconds \cite{jouppi2023tpu}.

\subsection{Training Datasets}
GPT-4V is a visual extension \cite{gpt4visioncard} of GPT-4 and has been trained with bulk on online image data along with various sources, including books, journals, code, and other text formats, which were included in the GPT-4 training dataset \cite{gpt4technical}. Thanks to intensive training, the AI model can now process and produce language and images that nearly match human-generated stuff. Because it is a generative model, GPT-4V can create new content in response to image and textual inputs. It can quickly and effectively analyze and understand the image using the inbuilt image decoding modules and massive amounts of textual data because of its dual-transformer design. It is particularly good at translating languages given in an image, for example, making cross-language communication easily written in an image. It may also produce various artistic content types and offer educational answers to user inquiries.

On the other hand, Gemini models are trained on a dataset that is both multimodal and multilingual. Their pertaining dataset uses data from web documents, books, and code and includes image, audio, and video data. They use the SentencePiece tokenizer \cite{kudo2018sentencepiece} and find that training the tokenizer on a large sample of the entire training corpus improves the inferred vocabulary and subsequently improves model performance.

\subsection{Performance}
Regarding performance, GPT-4V has shown significant improvements over previous models in visual input tasks. It has been reported to be more adept at generating factual responses and less likely to produce disallowed content, making it a more reliable choice in sensitive applications \cite{yang2023dawn,zhou2023exploring,gpt4visioncard}. In education, researchers found that GPT-4V is capable of understanding scoring rubrics and scoring students' drawn models to science phenomena with a certain degree of accuracy \cite{lee2023nerif}. 

With its sophisticated multimodal reasoning capabilities, Gemini understands and synthesizes insights from various inputs, including complex mathematical and scientific data \cite{gemini}. Google claimed that the newly released Gemini models notably achieved a milestone by reaching human-expert performance on the widely researched massive multitask language understanding exam benchmark. Furthermore, it enhanced the SOTA across all of the 20 multimodal benchmarks we investigated \cite{gemini}. 

However, the outcomes from Gemini Pro are not consistent. For example, Google's report compared the performance of Gemini with GPT-4V for their ability to understand images. They evaluated the models on four capabilities: high-level object recognition using captioning, fine-grained transcription using tasks, chart understanding, and multimodal reasoning using tasks \cite{akter2023depth}. The model is instructed to provide short answers aligned with the specific benchmark for zero-shot QA evaluation. All numbers are obtained using greedy sampling and without external optical character recognition tools. Results reported in Google's technical report have shown that GPT-4V outperformed Gemini Pro for all benchmarks with, on average, 7\% higher accuracy. We also have presented the results in Table~\ref{tab:image_understanding_comparison} by leveraging Google's permission to replicate and use the plots for research purposes. However, it is widely acknowledged that the capabilities of AI models, such as Gemini Pro and GPT-4V, in analyzing multimodal inputs and reasoning depend on context-specific tasks, which can provide a comprehensive qualitative explanation of their performance. Hence, in this study, we aim to analyze the performance of Gemini Pro and GPT-4V for educational tasks by providing assessment items, examples, rationales, and student responses in a condensed image and prompt models to categorize student responses based on the rationale provided in examples.

\begin{table}[h!]
\centering
\caption{Image understanding comparison between Gemini Pro and GPT-4V (Results reorganized from \citep{gemini})}
\label{tab:image_understanding_comparison}
\begin{tabular}{lcc}
\toprule
\textbf{Task} & \textbf{Gemini Pro} & \textbf{GPT-4V} \\
\midrule
MMMU (val) & 47.9\% & \textbf{56.8\%} \\
TextVQA (val) & 74.6\% & \textbf{78.0\%} \\
DocVQA (test) & 88.1\% & \textbf{88.4\%} \\
ChartQA (test) & 74.1\% & \textbf{78.5\%} \\
InfographicVQA (test) & 75.2\% & \textbf{75.3\%} \\
MathVista (testmini) & 45.2\% & \textbf{49.9\%} \\
AIZ2D (test) & 73.9\% & \textbf{78.2\%} \\
VQA v2 (test-dev) & 71.2\% & \textbf{77.2\%} \\
\bottomrule
\end{tabular}
\end{table}

\subsection{Capabilities}
GPT-4V's capabilities extend to processing both text and image inputs, making it versatile in applications like content creation, language translation, and educational tools. It has been integrated into various platforms, demonstrating its adaptability in different use cases \cite{gpt4visiontechnical,lyu2023gpt}. Gemini Pro, with its ability to handle a wider range of input types, is designed for both heavy-duty cloud applications and on-device solutions. This flexibility indicates a focus on scalability and accessibility in various environments \cite{gemini}.

\subsection{Safety}
Safety and alignment are key concerns in AI development, and both OpenAI and Google's Gemini team highlight the concerns on their website. GPT-4V has shown a significant reduction in its likelihood to respond to disallowed content requests and an increase in producing factual responses. These improvements are crucial for maintaining ethical standards in AI interactions \cite{gpt4visioncard} and reducing potential AI biases \citep{latif2023ai}. Gemini Pro has undergone extensive bias and toxicity analysis, with Google collaborating with external experts to identify and mitigate potential risks. Such efforts indicate the increasing importance placed on the ethical development of AI \cite{gemini}.

\subsection{Applications in Education}
Since its release, GPT-4V has been broadly integrated with existing technologies, ranging from integrating with Microsoft Bing \citep{yang2023dawn} for enhanced search capabilities to collaborating with Duolingo for language learning advancements \citep{gimpel2023unlocking}. These partnerships demonstrate GPT-4V's utility in improving user experience and knowledge management across domains \cite{wu2023visual}. 

\citep{senkaiahliyan2023gpt} demonstrated the capabilities of GPT-4V for clinical education through medical image interpretation and found that GPT-4V can identify and explain medical images but cannot provide safe clinical decisions and diagnostics. Similarly, \citeN{xu2023evaluation} evaluated GPT4-V's capabilities for ophthalmological studies and reported 63\% accuracy of GPT4-V in diagnosing ocular images. For automatic scoring by providing problem image and textual context with rubric, GPT-4V was able to achieve 51\% accuracy for science based assessments \citep{lee2023nerif}. Additionally, a comprehensive survey on multi-modality of AI for Education \citep{lee2023multimodality} focused on GPT-4V's capabilities to revolutionize education technology and challenges of stepping forward to artificial general intelligence.

Likewise, Gemini Pro, being integrated into products like Google Bard and Pixel, enhances reasoning, planning, and writing capabilities. Its availability through the Gemini API\footnote{\url{https://ai.google.dev/docs}} makes it a valuable resource for developers and enterprise customers, showcasing its potential applicability in education technology \cite{gemini}. Furthermore, \citeN{akter2023depth} reported Gemini-Pro's capabilities for solving math problems with a high accuracy of 69.67\% for the GSM8K dataset \citep{cobbe2021training}, but we did not find any such study as of December 23rd, 2023 on science educational assessment which again emphasize the significance of our study.

Overall, both GPT-4V and Gemini Pro represent significant advancements in the field of AI and language models. While they share some commonalities in terms of their multimodal capabilities, their differences in architecture, training methodologies, performance, safety measures, and applications illustrate AI technology's diverse and evolving landscape. A comprehensive comparison details are also presented in Table~\ref{tab:comprehensive_feature_comparison}. These models push the boundaries of what AI can achieve and raise important considerations for their ethical and practical implementation in various sectors. However, GPT4-V has shown higher image understanding performance than Gemini-pro, as evidenced by Table.~\ref{tab:image_understanding_comparison}.
\begin{table*}[h!]
\centering
\caption{Comprehensive Feature Comparison of Gemini-Pro and GPT-4V}
\label{tab:comprehensive_feature_comparison}
\begin{tabular}{lp{6.5cm}p{6.5cm}}
\toprule
\textbf{Feature} & \textbf{Gemini-Pro} & \textbf{GPT-4V} \\
\midrule
Model Type & Decoder-only Transformer & Autoregressive Transformer \\
Parameter Size & 280 billion & 1000 billion \\
Training Data & Google's internal datasets & OpenAI's publicly available datasets \\
Inference Hardware & TPUs (Tensor Processing Units) & GPUs (Graphics Processing Units) \\
Context Length & Supports 32K tokens & Supports up to 8K tokens \\
Safety Measures & Extensive bias and toxicity analysis & Improved safety and alignment over predecessors \\
Performance & Sophisticated multimodal reasoning & Advanced reasoning and problem-solving \\
Applications & Integrated into Bard, Pixel, and accessible via API & Integrated with Microsoft Bing, Duolingo, etc. \\
Real-world feedback & Rigorous testing with external experts & Continuous improvements based on user feedback \\
Multimodal Capabilities & Image, audio, video, and text & Text and image inputs \\
Fine-Tuning Capabilities & Tailored versions for different platforms & Fine-tuning with RLHF techniques \\
Applications in Education & Online article finding and summarization for research, Domain-specific content extraction from internet  & Automatic Scoring, Feedback system, paper writing, enhance creativity \\
\bottomrule
\end{tabular}
\end{table*}

\section{Methods} \label{methods} 

\subsection{Materials and Dataset}
This study reanalyzed student-created scientific models from a dataset derived from a primary study \cite{zhai2022applying}. The items, formulated by the NGSA team \citep{Harris2024Creating}, are designed to align with the \textit{NGSS} \citep{ngss2013next} performance expectations. They are part of a three-dimensional assessment strategy, integrating disciplinary core ideas, cross-cutting concepts, and science and engineering practices.

In our experiment involving six items, we selected 100 test cases for each item, using random sampling. The test datasets maintained a balanced distribution across three proficiency levels: 34 cases for 'Proficient,' 33 for 'Developing,' and 33 for 'Beginning,' with each category constituting one-third of the cases.

\subsection{Experimental Design}

This study conducted three experiments to answer the corresponding research questions. First, in the quantitative study, we compared the image classification performance of GPT-4V and Gemini Pro for automatic scoring of student-drawn models for science phenomena. We provided the two VLMs with the same prompt and image input for each task during this experiment. The performance metrics are quantitatively reported. Second, in the qualitative study, we heuristically explored the prompt that increases Gemini Pro's image classification performance. Third, we downsized the image input and examined how the performance of Gemini Pro changes.

\subsection{Prompt Design}\label{introduce NERIF}

For the prompt design, we adopted the NERIF (Notation-Enhanced Rubric Instruction for Few-shot Learning) method from \citeN{lee2023nerif}. In the NERIF, a user first writes a prompt with components essential to the task. After that, validation cases are used to confirm whether the prompt achieves the user intended for the task. If not, the \textit{Notation-Enhanced Scoring Rubric} is introduced/revised in the prompt, which combines \textit{human experts' scoring rules}, \textit{scoring rules aligned with proficiency levels}, and \textit{instructional notes} for better scoring. The validation and revision of the prompt are repeated until the improvement of the machine's performance reaches saturation. 

\begin{figure*}
    \centering
    \includegraphics[width=1\linewidth]{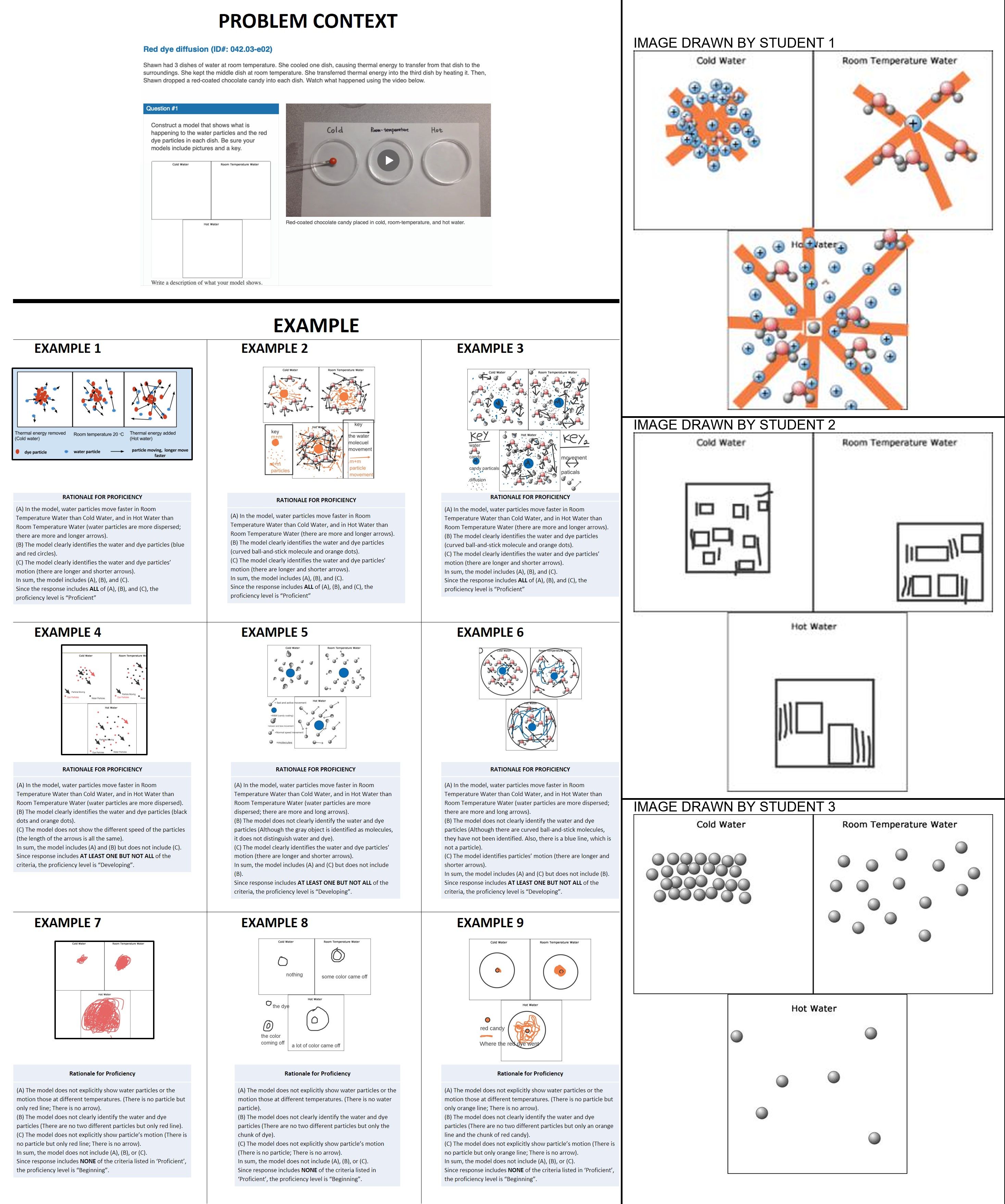}
    \caption{Example Input Image from Task 42}
    \label{fig:task42_agg_example}
\end{figure*}

\begin{figure*}
    \centering
    \includegraphics[width=1\linewidth]{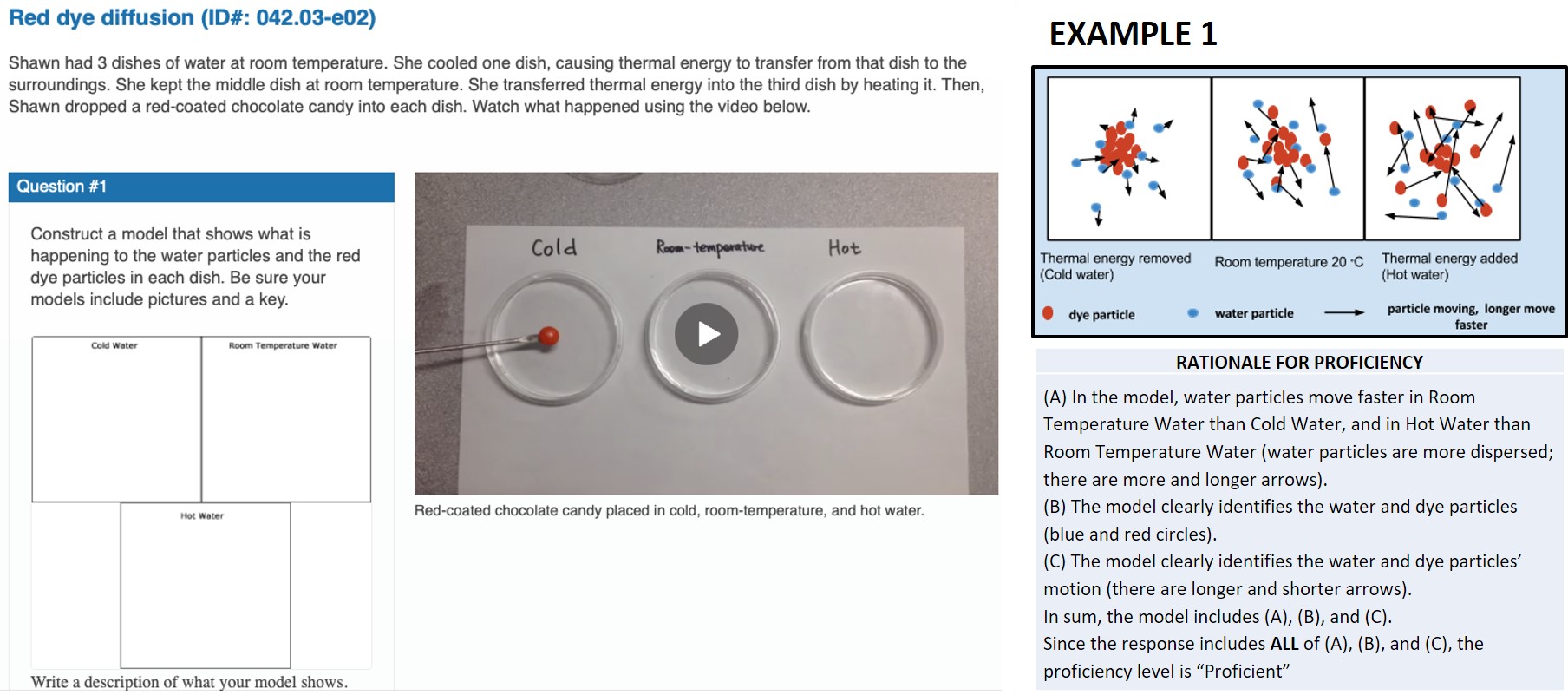}
    \caption{Example of Problem Context and Example from Task 42}
    \label{fig:task42_problem_example}
\end{figure*}

\begin{figure*}
    \centering
    \includegraphics[width=1\linewidth]{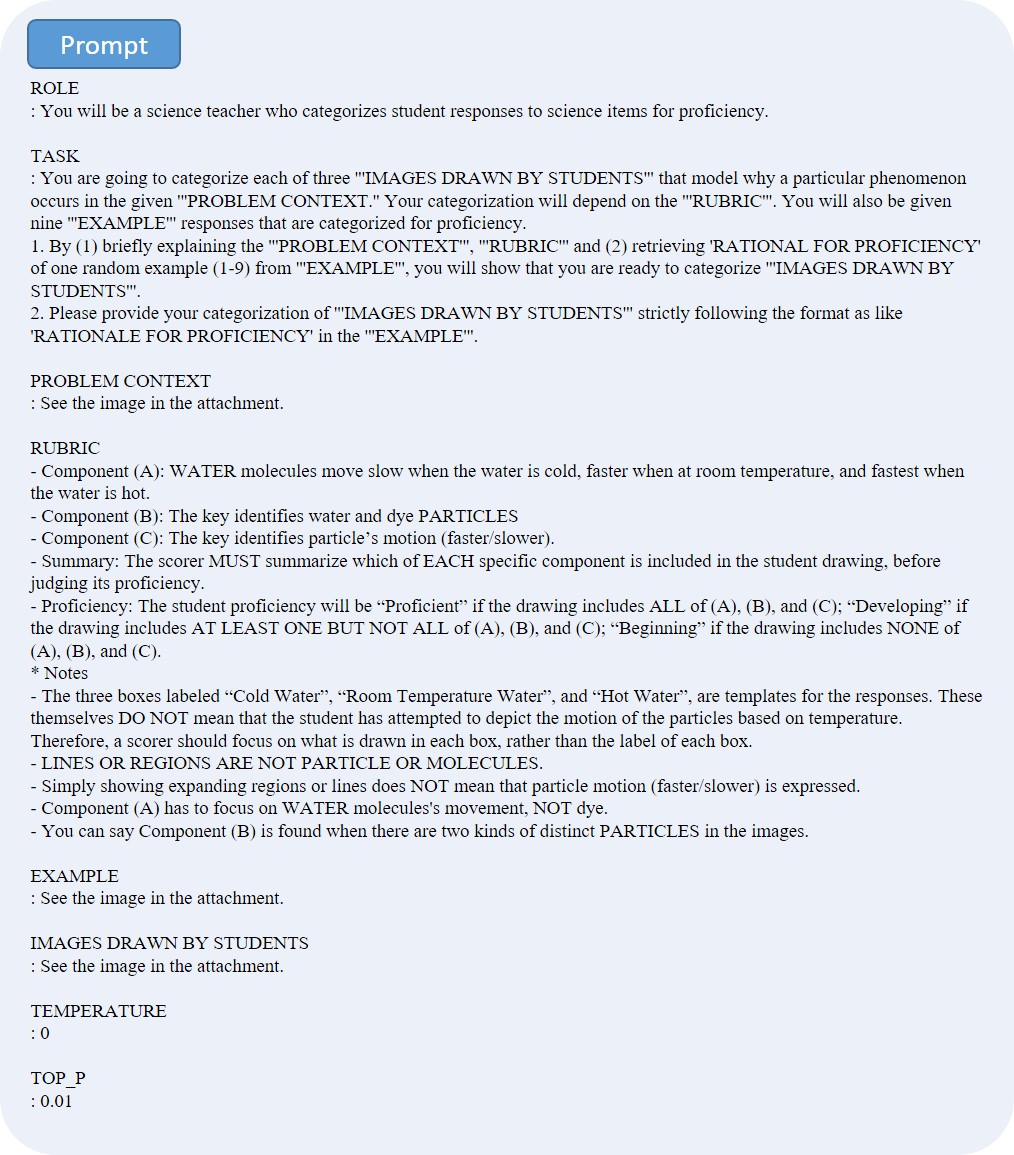}
    \caption{Example Prompt from Task 42}
    \label{fig:task42_prompt}
\end{figure*}

For the first experiment, we employed the prompt used in \citeN{lee2023nerif}, which used GPT-4V to analyze combined visual and text questions for automatic scoring, with slight revision. The input image and prompt used in \citeN{lee2023nerif} consisted of seven components, as shown in Figs. \ref{fig:task42_agg_example}-\ref{fig:task42_prompt}: (1) \textit{Role} that designates ChatGPT's role as a science teacher that scores student-drawn model, (2) \textit{Task} that explains what ChatGPT is requested to do, (3) \textit{Problem context} that ChatGPT has to retrieve from an image, (4) \textit{Notation-Enhanced Scoring Rubrics}, (5) Nine human scoring \textit{Examples} for few-shot learning (3 for 'Proficient,' 3 for 'Developing,' and 3 for 'Beginning' cases), (6) \textit{Models drawn by students} as test cases, and (7) Temperature/top\_p = 0/0.01 as hyper-parameters. In \citeN{lee2023nerif}, (3) and (5) were given in the first attached image (left of Fig. \ref{fig:task42_agg_example}; Fig. \ref{fig:task42_problem_example}), (6) in the second attached image (right of Fig. \ref{fig:task42_agg_example}), and others in the text (Fig. \ref{fig:task42_prompt})

The prompt has been slightly changed in this study, since there was an additional constraint as Gemini Pro was limited to processing up to one image as input, different from GPT-4V, which could process up to four drawn models simultaneously (as of Dec 12, 2023). To accommodate both LMMs, we changed the input protocol by merging (3), (5), and (6) into one image (Fig. \ref{fig:task42_agg_example}). The structure of image input and text prompt was consistent throughout the Tasks. The individual image given to the two VLMs was 2,935 (width) × 3,515 (height) pixels and about 1MB size large, throughout the tasks.

For the second experiment, we started from asking the VLMs "what do you see in the given image?" with the input image. After that, we asked them to "tell me about how the 'PROBLEM CONTEXT' is given in the attached image, in detail" to check whether they appropriately retrieve information from the image.

For the third experiment, in the situations where VLM(s) fail to proceed with the given Tasks (in the first experiment), we tried various strategies to make these done with the VLM(s), including breaking down the images and text prompt into smaller compartments.

\section{Findings} \label{findings}


 The details of the two models' scoring outcomes and potential reasons are presented in the following subsections.

\subsection{Scoring Performance on Student Drawn Models: Gemini Pro vs. GPT-4V}
\subsubsection{Scoring Accuracy}
We found that the response patterns of GPT-4V and Gemini Pro differed substantially. GPT-4V returned image classification results for all six Tasks. However, Gemini Pro produced image-classifying responses as instructed only for Task 42. For other Tasks, Gemini Pro produced various alternative forms of responses, rather than classification. Therefore, GPT-4V successfully scored students' drawn models  for 600 cases, while Gemini Pro only scored around 100 cases. This indicates Gemini Pro's incapability of processing a large image aggregated with both text prompts and image information.  The scoring performance of the two VLMs is presented in Table \ref{tab:Performance_Metrics}. Note that GPT-4V's performance is presented for every Task - in contrast, Gemini Pro's performance is presented only for Task 42, and all others are presented as 'NA' (Non-Available).

\begin{table*}[htp]
\centering
\caption{Image Classification Performance of GPT-4V and Gemini Pro on the NERIF Tasks}
\label{tab:Performance_Metrics}
\begin{tabular}{llllllllll}
\hline
\textbf{LLM} &
  \textbf{Task} &
  \textbf{Accuracy} &
  \textbf{Precision} &
  \textbf{Recall} &
  \textbf{F1} &
  \textbf{KappaQW}&
  \textbf{Acc\_Beg}&
  \textbf{Acc\_Dev} &
  \textbf{Acc\_Prof}\\ \hline
\textbf{GPT-4V}     & 42            & 0.47 & 0.50 & 0.47 & 0.46 & 0.37  & 0.62 & 0.56 & 0.24 \\
\textbf{}           & 44            & 0.44 & 0.60 & 0.44 & 0.38 & 0.26  & 0.88 & 0.31 & 0.12 \\
\textbf{} &
  45 &
  0.57&
  0.59&
  0.57&
  0.55&
  0.44&
  0.82&
  0.56&
  0.32\\
\textbf{}           & 48            & 0.41 & 0.35 & 0.41 & 0.36 & 0.29  & 0.65 & 0.56 & 0.03 \\
\textbf{}           & 53            & 0.46 & 0.53 & 0.46 & 0.45 & 0.38  & 0.56 & 0.63 & 0.21 \\
\textbf{}           & 57            & 0.54 & 0.57 & 0.54 & 0.54 & 0.50  & 0.65 & 0.59 & 0.38 \\
                    & \textbf{Mean} & 0.48 & 0.50 & 0.46 & 0.43 & 0.37  & 0.67 & 0.58 & 0.26 \\
\textbf{}           & \textbf{SD}   & 0.06 & 0.09 & 0.05 & 0.06 & 0.09  & 0.12 & 0.16 & 0.19 \\ \hline
\textbf{Gemini Pro} & 42            & 0.30 & 0.30 & 0.30 & 0.30 & -0.14 & 0.21 & 0.44 & 0.26 \\
\textbf{}           & 44            & NA   & NA   & NA   & NA   & NA    & NA   & NA   & NA   \\
\textbf{}           & 45            & NA   & NA   & NA   & NA   & NA    & NA   & NA   & NA   \\
\textbf{}           & 48            & NA   & NA   & NA   & NA   & NA    & NA   & NA   & NA   \\
\textbf{}           & 53            & NA   & NA   & NA   & NA   & NA    & NA   & NA   & NA   \\
\textbf{}           & 57            & NA   & NA   & NA   & NA   & NA    & NA   & NA   & NA   \\
                    & \textbf{Mean} & 0.3& 0.3& 0.30 & 0.3& -0.14 & 0.21 & 0.44 & 0.26 \\
                    & \textbf{SD}   & NA   & NA   & NA   & NA   & NA    & NA   & NA   & NA   \\ \hline
\end{tabular}
\end{table*}

Specifically, the mean accuracy of GPT-4V on the image classification Tasks was M = .48, SD = .06. Also, the mean precision was M = .50 (SD = .09), recall M = .46 (SD = .05), and F1 M =.43 (SD = .06). The category-wise accuracy was highest for 'Beginning' cases (M = .67; SD = .12), followed by 'Developing' (M = .58; SD = .16) and 'Proficient' cases (M = .26, SD = .19).

For Gemini Pro, we only received the scoring accuracy for Task 42. The mean accuracy was M = .3 (SD = NA). Also, the average precision was M =.3 (SD = NA), recall .30 (SD = NA), and F1 .3 (SD = NA). Note that the accuracy was less than .33, the expected value of random response in a trinomial classification task. The category-wise accuracy was highest for 'Developing' cases (M = .44), followed by 'Proficient' (M = .26) and 'Beginning' cases (M = .21).

To sum up, the number of successful production of the anticipated answer type (600 for GPT-4V versus 100 for Gemini Pro), and the classification accuracy (.48 for GPT-4V and .30 for Gemini Pro, which means GPT-4V shows 60\% higher accuracy than Gemini Pro) quantitatively show that GPT-4V's VQA performance on automatic scoring task is superior than that of Gemini Pro.

\subsubsection{Quadratic Weighted Cohen's Kappa}

Although scoring accuracy provides a measure to understand the performance of Gemini Pro on understanding the scoring rubric and automatic scoring of drawn models compared to GPT-4V, this measure does not reflect that machine-human disagreements differ, some disagreement are more severe than others. For example, misscoring a "Proficient"-level student-drawn model as "Beginning" is more severe than as "Developing," and thus should receive more penalty. To further understand this difference between the two VLMs, we compared the confusion matrices of automatic scoring for Task 42 between Gemini Pro and GPT-4V (Table \ref{tab:confusion_42}) and calculated the Quadratic Weighted Cohen's Kappa,

\[
\kappa = \frac{P_o - P_e}{1 - P_e}
\]

where

\[
P_o = 1 - \frac{\sum_{i=1}^{k}\sum_{j=1}^{k}w_{ij}x_{ij}}{N}
\]

and

\[
P_e = 1 - \frac{\sum_{i=1}^{k}\sum_{j=1}^{k}w_{ij}e_{ij}}{N^2}
\]

In these formulas:

\begin{itemize}
    \item $k$ is the number of rating categories.
    \item $w_{ij} = (i - j)^2$ is the weight for the disagreement between categories $i$ and $j$, representing quadratic weighting.
    \item $x_{ij}$ is the observed count of ratings in the cell corresponding to category $i$ by rater 1 and category $j$ by rater 2.
    \item $e_{ij}$ is the expected count under chance agreement, calculated as $(\text{row total of } i \times \text{column total of } j) / N$.
    \item $N$ is the total number of ratings.
\end{itemize}

As presented in Table \ref{tab:Performance_Metrics}, the Quadratic Weighted Kappa of GPT-4V on Task 42 was .37. And that for other tasks spanned from .26 to .50, with M = .37 and SD = .09, which can be considered as 'Fair' to 'Moderate' level \citep{LandisKoch1977}.  In contrast, the Quadratic Weighted Kappa of Gemini Pro on Task 42 was -.14. Note that since Kappa aims to correct chance agreement, value 0 indicates that all agreements are by chance. A negative value suggests that the agreement is worse than guessing. This could be because Gemini misinterpreted the information and made systematically. (see Table \ref{tab:confusion_42}




\begin{table}[]
\centering
\caption{Confusion Matrices of GPT-4V and Gemini Pro on Task 42}
\label{tab:confusion_42}
\resizebox{\linewidth}{!}{\begin{tabular}{l|ccc|ccc}

\hline
\textbf{True Label} & \multicolumn{3}{c}{\textbf{GPT-4V's Prediction}}          & \multicolumn{3}{|c}{\textbf{Gemini Pro's Prediction}}      \\
\hline
           & Beginning & Developing & Proficient & Beginning & Developing & Proficient \\
Beginning  & 21        & 13         & 0          & 7         & 14         & 13         \\
Developing & 8         & 18         & 6          & 11        & 14         & 7          \\
Proficient & 7         & 19         & 8          & 12        & 13         & 9          \\ \hline
\end{tabular}}
\end{table}



To sum up, the scoring accuracy, quadratic weighted Cohen's Kappa and confusion matrix show that GPT-4V's image processing capability for automatic scoring is superior to Gemini Pro.

\subsection{Qualitative Characteristics of Scoring by Gemini Pro vs. GPT-4V}

To uncover the divergent performance between Gemini Pro and GPT-4V, we qualitatively analyzed the scoring patterns of GPT-4V and Gemini Pro with the NERIF prompting methods. Note that we not only present what the two LLM's returned for the complete prompt and input image, but also the results of heuristic examination of their VQA performance. 

\subsubsection{Finding 1: Gemini Pro fails to recognize fine-grained texts printed in the image | GPT-4V succeeds}

\begin{figure*}
    \centering
    \includegraphics[width=1\linewidth]{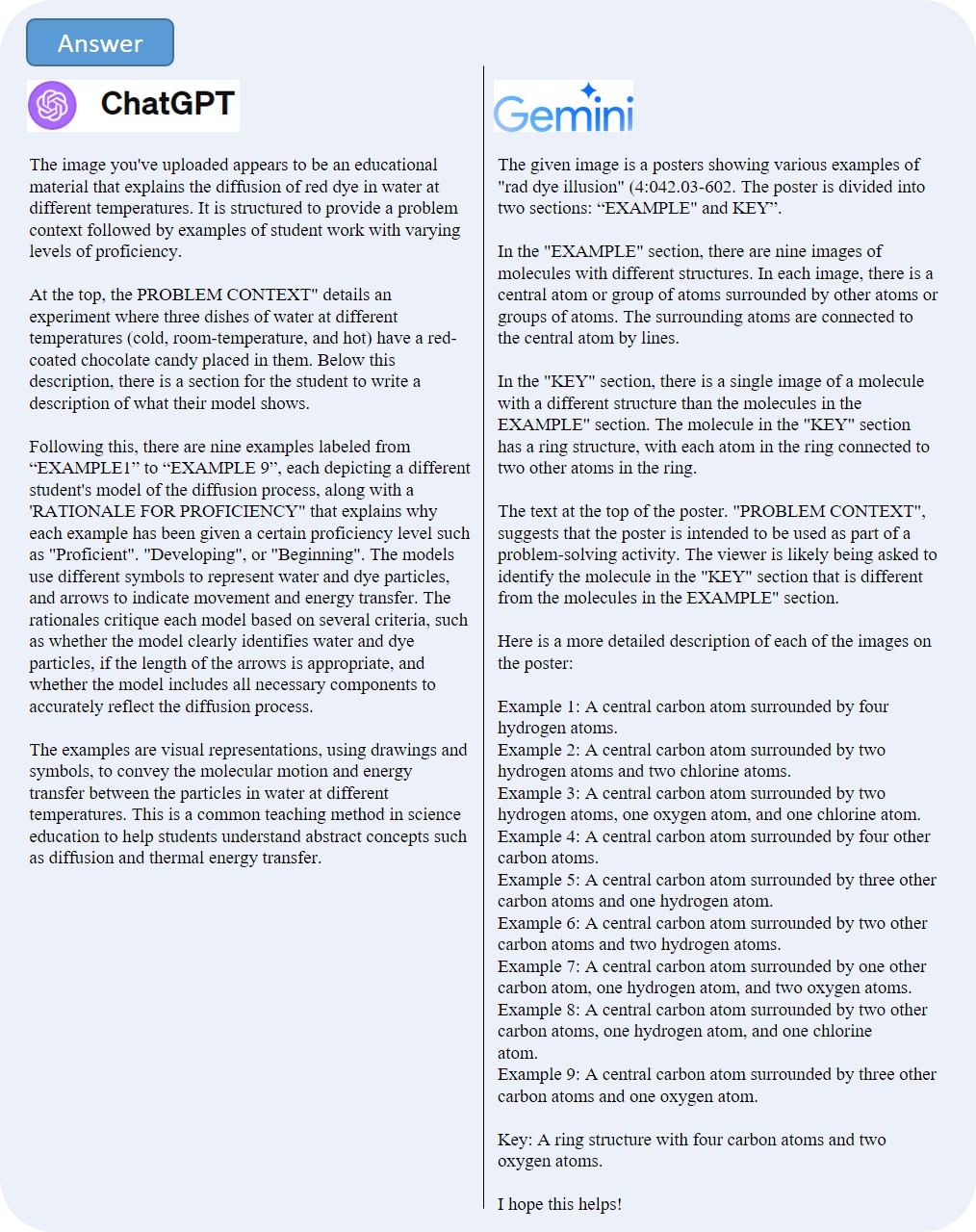}
    \caption{Answers from GPT-4V and Gemini Pro When Asked "What Do You See in the Given Image?" (Left: GPT-4V; Right: Gemini Pro)}
    \label{fig:qualitative_1}
\end{figure*}

We gave the problem context and example (left of Fig. \ref{fig:task42_agg_example}) as one input image and asked the two VLMs, "What do you see in the given image?". The outcomes from GPT-4V and Gemini Pro are presented in Fig. \ref{fig:qualitative_1}. GPT-4V responded, "The image you've uploaded appears to be an educational material that explains the diffusion of red dye water at different temperatures. ..." which is correct information about the given image (left of Fig. \ref{fig:qualitative_1}). It also correctly described the locations and details of the problem context, nine examples, rationale for proficiency, three proficiency levels, etc. In contrast, Gemini Pro responded that "The given image is a poster showing various examples of a "rad dye illusion" (4:042.03-602). The poster is divided into two sections: "EXAMPLE" and "KEY"." (right of Fig. \ref{fig:qualitative_1}). This falsely describes the information in the given image. The image is not a poster; The image says about "Red dye diffusion (ID\#: 042.03-e02)", not "Rad dye illusion (4:042.03-602)". Also, the image is divided into "PROBLEM CONTEXT" and "EXAMPLE," not "EXAMPLE" and "KEY."

This outcome shows that Gemini Pro fails to precisely recognize fine-grained texts in the image and generates false information, while GPT-4V succeeds with the same image.

\subsubsection{Finding 2: When failing to recognize the image, Gemini Pro often considers the image as a scientific poster}

\begin{figure*}
    \centering
    \includegraphics[width=1\linewidth]{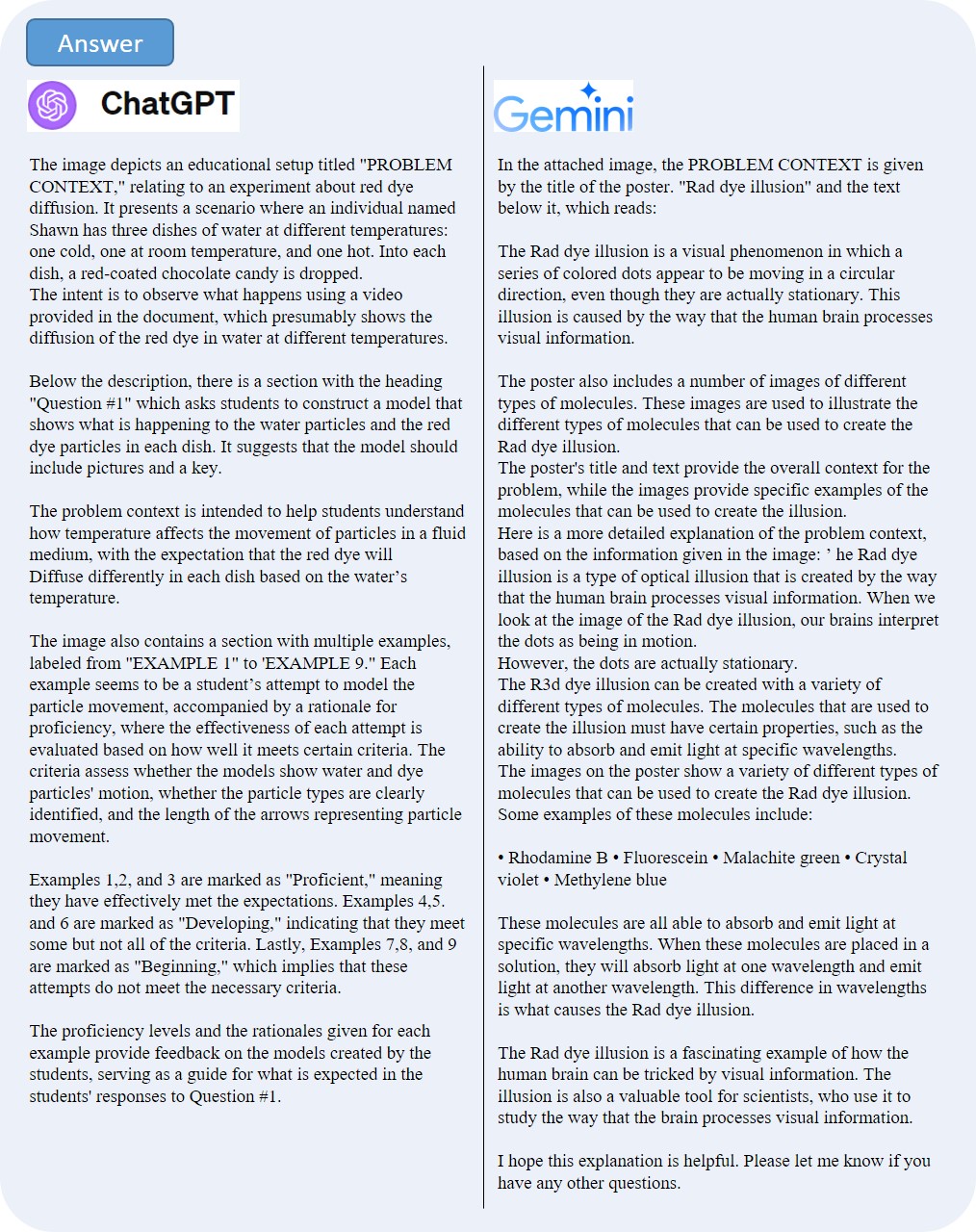}
    \caption{Scoring Outcomes from GPT-4V and Gemini Pro When Asked "Tell me about how the 'PROBLEM CONTEXT'" (Left: GPT-4V; Right: Gemini Pro)}
    \label{fig:qualitative_2}
\end{figure*}

We gave the same image as the above and asked the two VLMs, “Tell me about how the 'PROBLEM CONTEXT' is given in the attached image, in detail" to see whether they retrieve the information as requested by the user. The scoring outcomes from GPT-4V and Gemini Pro are presented in Fig. \ref{fig:qualitative_2}, which are similar to Fig. \ref{fig:qualitative_1}. GPT-4V correctly retrieved the problem context, an "experiment about red dye diffusion" conducted by "Shawn," in the situation where "the red dye will diffuse differently in each dish based on the water's temperature" (left of Fig. \ref{fig:qualitative_2}). In contrast, Gemini Pro returned, "the PROBLEM CONTEXT is given by the title of the poster, "Rad dye illusion".," which is wrong. Also, it fabricated non-existent information such as "the text ... reads: The Rad dye illusion is a visual phenomenon in which a series of colored dots appear to be moving in a circular direction, ...," and also falsely said that "the images on the poster show ... Rhodamine B, Fluorescein, Malachite green, ... ." It is noteworthy that Gemini Pro recognized the given image as a scientific poster and generated hallucinated information based on it (Figs. \ref{fig:qualitative_1}-\ref{fig:qualitative_2}). This implies that its training dataset could have included scientific posters.

\subsubsection{Finding 3: Gemini Pro fails to retrieve a random example from few-shot examples | GPT-4V succeeds}

\begin{figure*}
    \centering
    \includegraphics[width=1\linewidth]{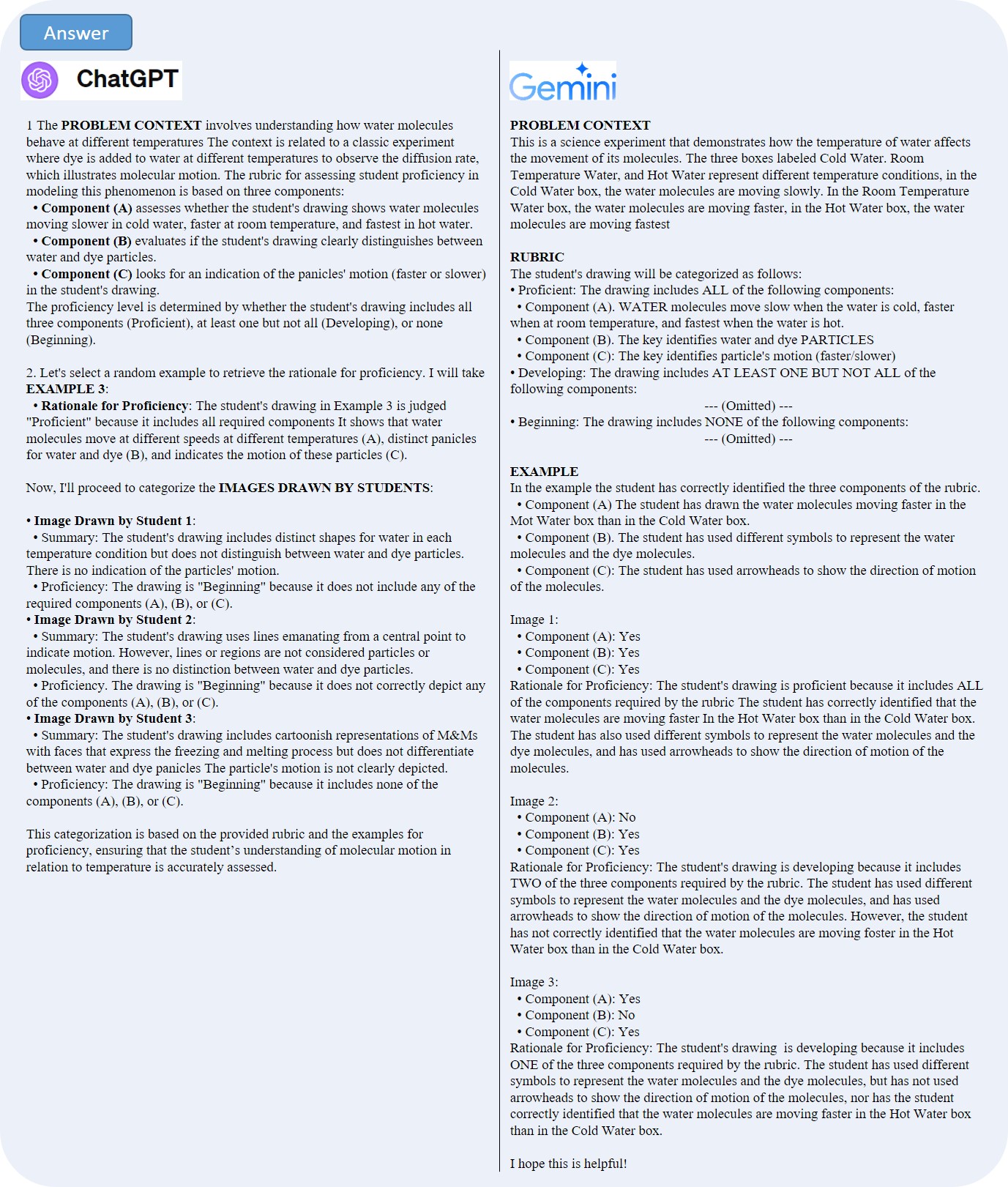}
    \caption{Scoring Outcomes from GPT-4V and Gemini Pro When Given the Full Image and Prompt Input (Left: GPT-4V; Right: Gemini Pro)}
    \label{fig:qualitative_3}
\end{figure*}

We gave GPT-4V and Gemini Pro an image for Task 42, similar to Fig. \ref{fig:task42_agg_example} with three 'Beginning' student-drawn models on the right side. The automatic scoring results of GPT-4V and Gemini Pro are presented in Fig. \ref{fig:qualitative_3}. Notably, GPT-4V strictly followed the instruction that requires it to choose a random one out of the nine examples, and correctly retrieved that "Example 3 is judged "Proficient"." It had further correctly retrieved in the image why example 3 is labeled as "Proficient." After that, it scored the three 'Beginning' examples correctly. In contrast, although Gemini Pro seems to say something about an example, it did not specify which example it retrieved, and thus there was no evidence it retrieved one. After that, it scored the three 'Beginning' examples as 'Proficient,' 'Developing,' and 'Developing,' which were all mis-scored. \citeN{lee2023nerif} experimentally reported that GPT-4V is likely to score student-drawn models when it had been provided with few-shot examples and retrieved one of them, which explains how GPT-4V succeeded in this qualitative case. If this is also the case for Gemini Pro, its failure to retrieve a certain example from the image input could be one of the reasons for its low scoring accuracy.

\subsection{Limited Improvement of Gemini Pro's Performance on the Automatic Scoring Task}

\begin{figure*}
    \centering
    \includegraphics[width=1\linewidth]{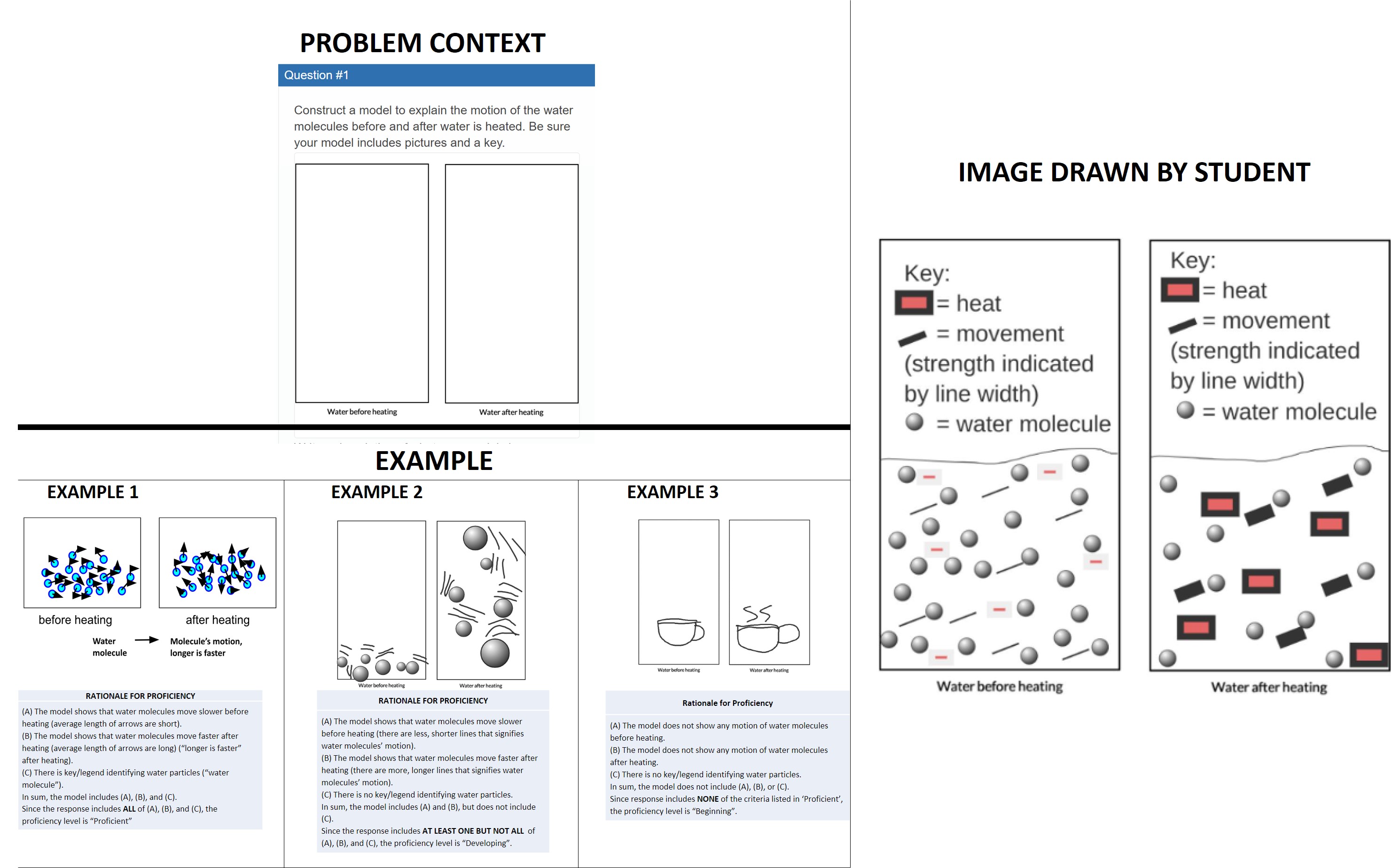}
    \caption{Example of Simplified Input Image from Task 53}
    \label{fig:qualitative_5}
\end{figure*}

\begin{figure*}
    \centering
    \includegraphics[width=1\linewidth]{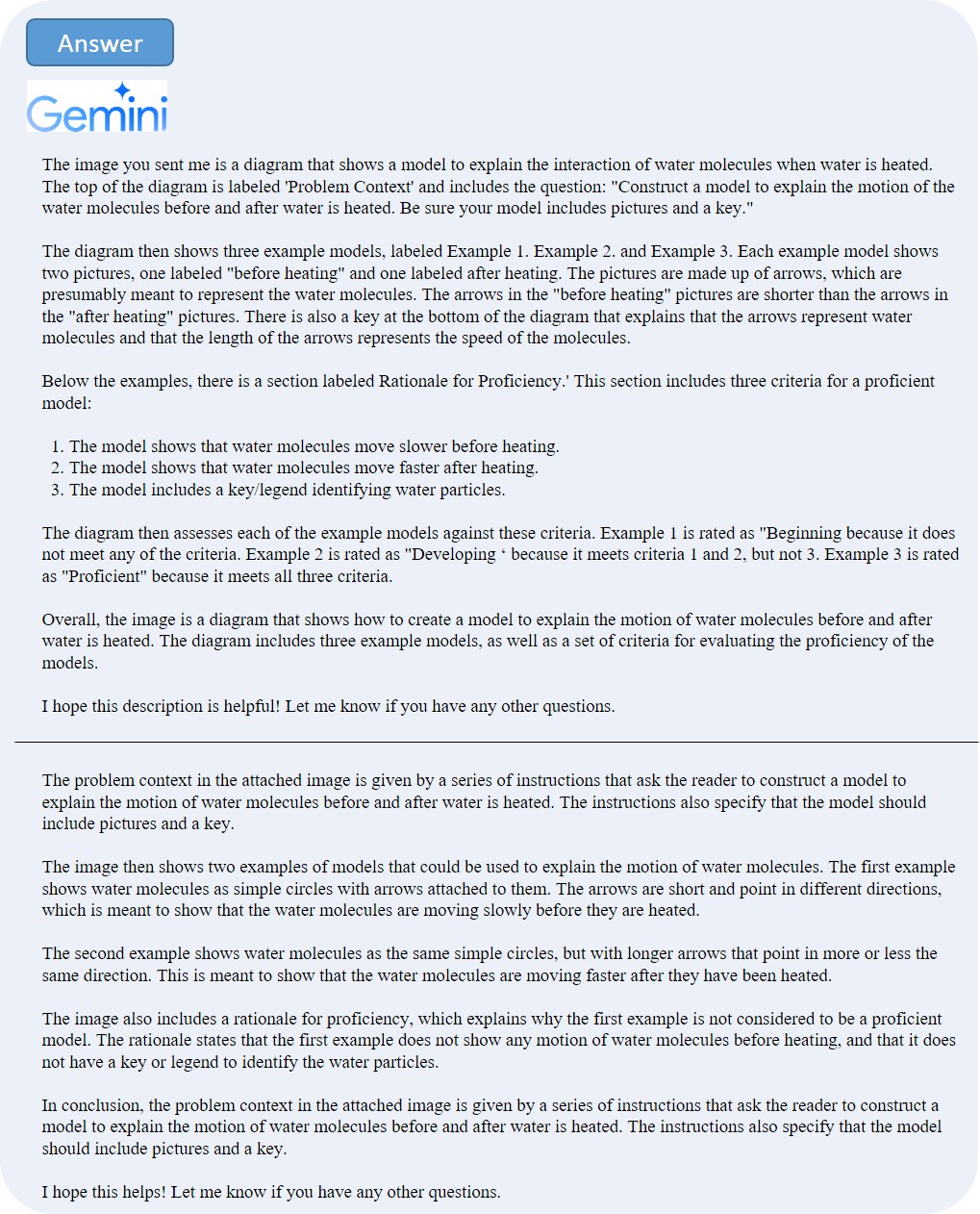}
    \caption{Scoring from Gemini Pro with Simplified Input Image (Top: "What do you see in the given image?" GPT-4V; Bottom: “Tell me about how the 'PROBLEM CONTEXT' is given in the attached image, in detail.")}
    \label{fig:qualitative_5_1}
\end{figure*}

Given that Gemini Pro could not process the fine-grained, large-size input image as well as GPT-4V, we broke down the input image to reduce the complexity of the input image prompt and expected to improve Gemini Pro's scoring performance. We applied this strategy to Task 53, one of the previously unsuccessful Tasks. We reduced the number of few-shot learning examples from nine to three (a set of one example for each of the 'Beginning,' 'Developing,' and 'Proficient' category; Fig. \ref{fig:qualitative_5}). We provided Gemini Pro with the input image without test cases (Left of Fig. \ref{fig:qualitative_5}) and asked "What do you see in the given image?" and “Tell me about how the 'PROBLEM CONTEXT' is given in the attached image, in detail." For the former question, it correctly responded, "The image you sent to me is ... to explain the interaction of water molecules when water is heated." (Top of Fig. \ref{fig:qualitative_5_1}). It successfully explained the location of the problem context, three examples, and 'Rationale for Proficiency." However, it responded that example 1 is a 'Beginning' and example 3 is 'Proficient,' which are both incorrect. For the latter question, it correctly stated that "the problem context in the attached image is given by series of instructions task ask the reader to construct a model ..." (Bottom of Fig. \ref{fig:qualitative_5_1})
. However, it incorrectly responded that "the image then shows two examples." This is even inconsistent with the very above answer, which correctly identified three examples given in the same input image. When we tried the same questions using an image with nine few-shot examples for comparison, Gemini Pro started malfunctioning and responded that the image is a "poster" again. This result implies that Gemini Pro's scoring performance was improved with less information in and less pixel size of the input image. However, it still failed to precisely retrieve information from the image. Also, when we tried to get automatic scoring results for one student-drawn image by providing it with Fig. \ref{fig:qualitative_5}, Gemini Pro did not return any scoring result. In summary, reducing the input image size improved Gemini Pro's outcome for retrieving information from the image, but this change was insufficient for automatic scoring tasks.


\section{Discussion} \label{discussion}
The study aimed to compare Gemini Pro and GPT-4V in educational settings, particularly in scoring student-drawn models using NERIF. Major findings highlighted GPT-4V's superior accuracy in image classification compared to Gemini Pro and its adeptness at processing detailed text in images, evidenced in scoring students' drawn models. The study also uncovered the nuance of Gemini's scoring performance, which accounts for the lower performance compared to GPT-4V, including failing to recognize fine-grained texts printed in the image, often considering the image as a scientific poster, and failing to retrieve a random example from few-shot examples. Even adapting NERIF for Gemini Pro was not improved enough to be comparable to GPT-4V. The findings contribute to the literature in several aspects.

First, the findings contribute significantly to the existing literature by showcasing GPT-4V's remarkable scoring accuracy, distinguishing it from Gemini Pro, and setting new benchmarks. This performance aligns with trends noted in the literature \citep{zhai2020applying}, which highlighted the evolving precision of AI in the automatic scoring of constructed response assessments. GPT-4V's ability to accurately score complex student-drawn models aligns with the findings of \cite{lee2023nerif}, who underscored the potential of AI in enhancing visual assessments. Furthermore, unlike earlier automatic scoring systems that primarily focused on textual data \citep{zhai2020from}, GPT-4V integrates advanced image processing, marking a significant leap in the capability of AI tools. This study not only corroborates the growing efficacy of AI in educational settings but also extends it by demonstrating practical applications in multimodal formats of student responses. Such advancements address some of the limitations discussed in earlier research and the challenges in accurately assessing non-textual student work \citep{zhai2022applying, lee2023automated}. By exploring the reasons behind these findings, the study contributes to a deeper understanding of AI's role in educational assessment, a field ripe for further exploration and innovation.

The findings of this study, highlighting the nuanced limitations of Gemini's scoring performance compared to GPT-4V, contribute notably to uncovering the mechanisms of the newly released Gemini Pro in image analysis. The findings overall seem inconsistent with the major report of \citeN{gemini}. Gemini's challenges in recognizing fine-grained text within images and its tendency to misclassify images as scientific posters resonate with concerns raised in earlier research. For instance, studies by \citeN{RN3456} pointed out similar difficulties faced by Gemini in discerning detailed textual elements in complex visual contexts.  Furthermore, Gemini's failure to effectively utilize few-shot visual examples reflects the limitations discussed by \citeN{fu2023challenger}, highlighting the challenges of interpreting images with a large number of elements because of its concision approach. This study's examination of these specific shortcomings not only corroborates the observations from previous research but also provides a concrete comparison of how different AI models handle complex educational tasks. Such insights are invaluable for the ongoing development of more sophisticated and context-aware AI tools in education, as suggested by the work of \citeN{Latif2023AGI}, who emphasized the need for AI systems to better adapt to the nuanced and varied nature of educational content and assessment methodologies.

The study's findings on the limitations of adapting NERIF for Gemini Pro, which did not significantly enhance its performance to match that of GPT-4V, offer several contributions to the existing literature on AI in educational contexts. This suggests the challenges of integrating specific AI frameworks into existing models, often finding that such adaptations do not always yield the expected improvements in performance. Furthermore, the results highlighted the complexity of AI systems in education, noting that modifications like NERIF require careful calibration to align with the intricacies of educational content. The inability of Gemini Pro to reach the accuracy level of GPT-4V, even with NERIF, underscores that AI advancements in education depend not just on the initial capabilities of the AI model but also the the intrinsic design of prompts. This study thereby adds a nuanced perspective to the discussion on the limits of augmenting AI systems with additional frameworks and suggests the need for a more holistic approach to AI development in educational settings. The findings highlight the intricate balance between AI adaptability and the inherent design of educational AI systems, contributing to a deeper understanding of how AI can be effectively tailored and utilized in complex educational assessments.

\section{Conclusions} \label{conclusions}

This study presented a comprehensive comparison between Gemini Pro and GPT-4V in educational settings, focusing on their ability to score student-drawn models using NERIF. The findings highlighted GPT-4V's superior accuracy in image classification and its proficiency in processing detailed text in images, demonstrating its potential for enhancing multimodal assessments in education. The study revealed that even with adaptations to NERIF, Gemini Pro could not match GPT-4V's performance, emphasizing the complexities in AI model adaptation and the importance of intrinsic design and initial capabilities. These results contribute significantly to the literature on AI in education, suggesting the need for more sophisticated, context-aware AI tools. This study adds to the understanding of AI's role in educational assessment, indicating directions for future research and development in this rapidly evolving field.

\begin{acks}
This study was funded by the National Science Foundation (NSF) (Award no. 2101104). Any opinions, findings, conclusions, or recommendations expressed in this material are those of the author(s) and do not necessarily reflect the views of the NSF. The authors thank the NGSA team and the researchers involved in the parental study (Zhai, He, and Krajcik, 2022) and those who coded the student-drawn models.
The authors specifically thank Joon Kum, who helped developing the prompts and making predictions on the test data.
\end{acks}

\bibliographystyle{ACM-Reference-Format}
\bibliography{sample-base}

\appendix

\end{document}